\documentclass[10pt,twocolumn,letterpaper]{article}

\usepackage{cvpr}
\usepackage{times}
\usepackage{epsfig}
\usepackage{graphicx}
\usepackage{amsmath}
\usepackage{amssymb}

\usepackage{bbm}
\usepackage{booktabs}
\usepackage{multirow}
\usepackage{array}
\usepackage{verbatim}

\usepackage[colorlinks,breaklinks=true,bookmarks=false]{hyperref}

\cvprfinalcopy 

\def\cvprPaperID{5409} 

\ifcvprfinal\fi
\begin{document}

\title{Robust Partial Matching for Person Search in the Wild}

\author{Yingji Zhong$^{1}$, Xiaoyu Wang$^{2,3}$, Shiliang Zhang$^{1}$\\
$^{1}$Peking University, $^{2}$Intellifusion, $^{3}$The Chinese University of Hong Kong, Shenzhen\\
{\tt\small $^{1}$\{zhongyj, slzhang.jdl\}@pku.edu.cn, $^{2,3}$fanghuaxue@gmail.com}
}

\maketitle
\thispagestyle{empty}

\begin{abstract}
Various factors like occlusions, backgrounds, \emph{etc}., would lead to misaligned detected bounding boxes , \emph{e.g.}, ones covering only portions of human body. This issue is common but overlooked by previous person search works. To alleviate this issue, this paper proposes an Align-to-Part Network (APNet) for person detection and re-Identification (reID). APNet refines detected bounding boxes to cover the estimated holistic body regions, from which discriminative part features can be extracted and aligned. Aligned part features naturally formulate reID as a partial feature matching procedure, where valid part features are selected for similarity computation, while part features on occluded or noisy regions are discarded. This design enhances the robustness of person search to real-world challenges with marginal computation overhead. This paper also contributes a Large-Scale dataset for Person Search in the wild (LSPS), which is by far the largest and the most challenging dataset for person search. Experiments show that APNet brings considerable performance improvement on LSPS. Meanwhile, it achieves competitive performance on existing person search benchmarks like CUHK-SYSU and PRW.
\end{abstract}

\section{Introduction}

Thanks to recent research efforts~\cite{wei2017glad,su2017pose,kalayeh2018human,hermans2017defense,wang2018mancs,sun2018beyond,yang2019towards,zheng2019re,zheng2018Pyramidal}, the performance of person re-Identification (reID) has been significantly improved. However, one potential issue with current reID setting is that, it treats person detection and reID as two isolated steps. This issue can be compensated by person search~\cite{tong2017Joint,zheng2017Person,chen2018Person,xu2018Person,han2019Re,munjal2019Query} to jointly accomplish person detection and reID. Being able to jointly optimize person detection and reID, person search is attracting more and more attention, and is potential to present advantages in flexibility, efficiency, and accuracy.

\begin{figure}[t]
\centering
\begin{center}
\includegraphics[width=0.46\textwidth]{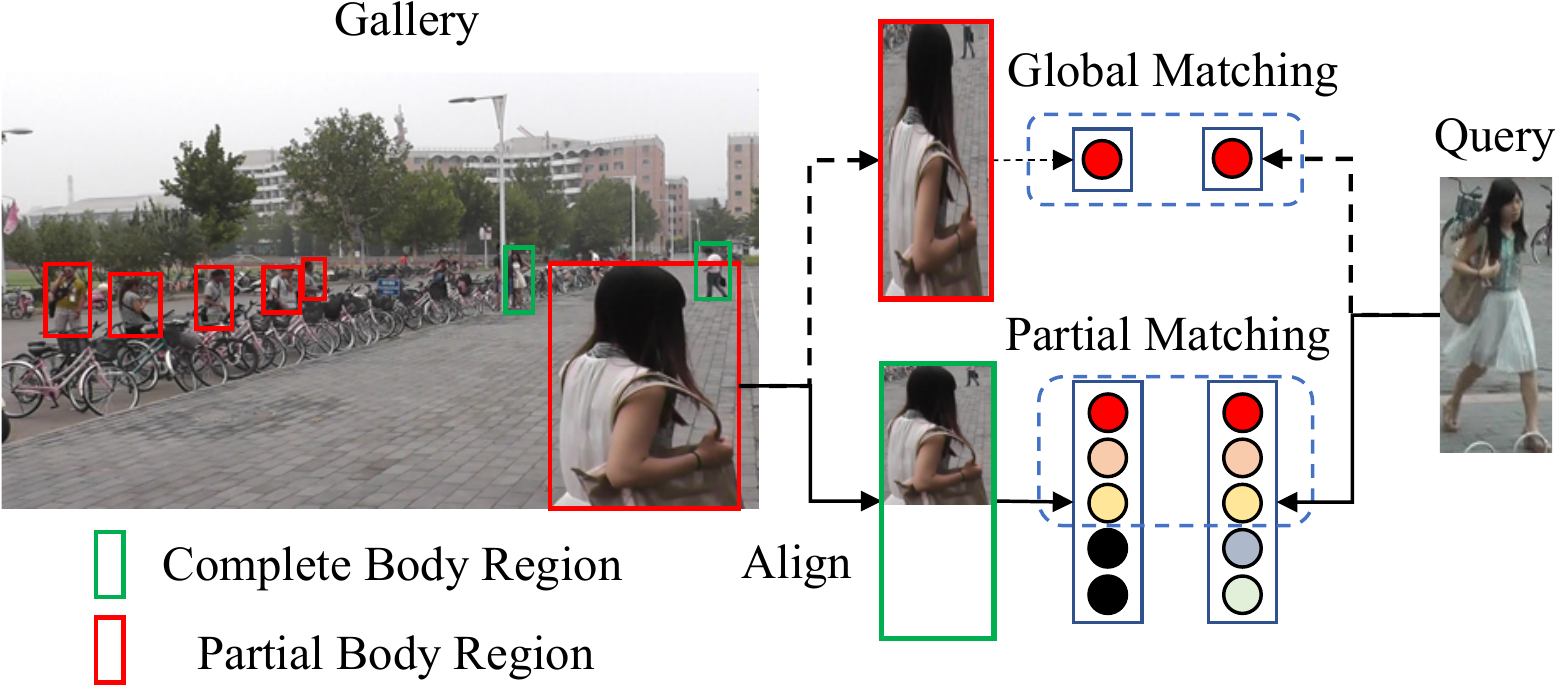}\\ 
\end{center}
\caption{Illustration of misaligned person bounding boxes and issues of person matching with global feature. Misalignments commonly occur in person search and can not be eliminated by training stronger detectors. This paper addresses this issue with bounding box alignment and partial matching.}
\label{fig:motivation}
\vspace{-0.2cm}
\end{figure}

Compared with reID, person search needs to design pedestrian detectors robust to variances of scales, illuminations, backgrounds, occlusions, \emph{etc}. Some person search works refine detectors by jointly training detectors and reID models~\cite{tong2017Joint,munjal2019Query,hao2017Neural,han2019Re}. Although strong detectors are generally helpful for reID, they may still produce misaligned person bounding boxes. As shown in Fig.~\ref{fig:motivation}, occlusions and limited camera viewing field lead to many accurate but misaligned bounding boxes covering portions of pedestrians. Most of existing person search methods extracts global feature from the detected bounding boxes, no matter the joint models~\cite{tong2017Joint,munjal2019Query} or the separate models~\cite{chen2018Person,xu2018Person,han2019Re}. As shown in Fig.~\ref{fig:motivation}, misalignment degrades the performance of global features, because it is not reasonable to match partial features against global features. More detailed reviews to person search works can be found in Sec.~\ref{sec:relatedwork}.

This paper targets to design a unified person search framework robust to the misaligned bounding boxes. As shown in Fig.~\ref{fig:motivation}, the basic idea is to refine detected bounding boxes to cover the estimated holistic body regions to eliminate misalignment errors. Then, aligned part features could be extracted from refined bounding boxes with simple region division. Part features hence allow for robust partial matching across bounding boxes, where features on the mutual visible regions are matched for similarity computation, and features on noisy regions are not considered for matching. In other words, our strategy improves both the feature resolution and robustness to misalignment errors.

The above idea leads to the Align-to-Part Network (APNet) for person search. Built on OIM~\cite{tong2017Joint}, APNet consists of the additional Bounding Box Aligner (BBA) and Region Sensitive Feature Extractor (RSFE). The BBA module is implemented by predicting 4 bounding box offset values. It can be trained by automatic data augmentation without requiring manual annotations. Because of enlarged receptive fields of neurons in feature extraction CNN layer, noisy or occluded parts would affect features of their adjacent parts in the same bounding box. We further design the RSFE module to extract part features. RSFE reinforces local cues in each part feature, hence effectively alleviates the negative effects of adjacent noises in part feature learning.

We test our approach on current person search benchmarks, \emph{i.e.}, PRW~\cite{zheng2017Person} and CUHK-SYSU~\cite{tong2017Joint}. Experiment results show that, our APNet achieves competitive performance compared with recent works. To test our approach in more challenging scenarios, we contribute a new person search dataset named as Large-Scale dataset for Person Search in the wild (LSPS). Compared with PRW and CUHK-SYSU, LSPS presents several new features that would encourage the research towards more realistic person search: 1) large scale: it contains a large number of pedestrians and bounding boxes; 2) automatically detected bounding boxes, which differ with the manually annotated ones in PRW and CUHK-SYSU; 3) collected in scenarios with occlusions and crowdings. On this challenging LSPS, our method exhibits substantial advantages.

Compared with reID, person search is still relatively under-explored. It is critical to deal with the challenge occuring in both person detection and reID. To the best of our knowledge, the proposed APNet is an early attempt in boosting the robustness to misalignment issue in person search. As shown in our experiments, the APNet brings considerable performance gains. The proposed LSPS dataset simulates realistic scenarios and presents a more challenging person search task than existing datasets. We believe that, the proposed methods and the contributed LSPS can inspire more research efforts in person search.

\section{Related Work} \label{sec:relatedwork}

This paper is closely related to person reID and person search. The following parts review those two lines of works.

\emph{Person reID} extracts robust and discriminative features from given person bounding boxes. Current fully supervised reID methods can be briefly summarized into the following categories. 1) Learn local features of regions utilizing offline detector to split body into several regions~\cite{wei2017glad,su2017pose,kalayeh2018human,li2019pose}, or uniformly split the body into several stripes~\cite{zheng2018Pyramidal,sun2018beyond,wang2018Learning}; 2) Learn the robust feature utilizing additional attribute annotation~\cite{su2016deep,su2015multi,su2017attributes,tay2019aanet}; 3) Enhance the discriminativeness of feature by attention mechanism~\cite{li2018harmonious,hou2019interaction,wang2018mancs}; 4) Impose constraints on feature space with loss functions, like verification loss~\cite{zheng2018discriminatively} and triplet loss~\cite{hermans2017defense}. Among the above methods, PCB~\cite{sun2018beyond} and its variants~\cite{zheng2018Pyramidal,wang2018Learning} dominate the state-of-the-art performance. PCB uniformly splits the feature map into stripes and supervises them to ensure their discriminative power. However, PCB requires a strict alignment of the input images since misalignment breaks the correspondence of stripes on the same spatial position. The performance of PCB thus degrades substantially when misalignments exist~\cite{sun2019Perceive}.

There are some reID works focusing on partial reID. Zheng \textit{et al.}~\cite{zheng2016Partial} propose AMC+SWM to solve the partial reID. AMC collects patches from all gallery images and set up a dictionary for patch level matching. He \textit{et al.}~\cite{he2018Deep} propose DSR to reconstruct the partial image by holistic image and utilize the reconstruction error as the similarity. Both AMC~\cite{zheng2016Partial} and DSR~\cite{he2018Deep} are optimisation-based thus it is expensive to get the similarity of each query-gallery pair. PGFA~\cite{miao2019pose} applies partial matching scheme with the aligned input images, whose occluded parts are indicated by the expensive offline keypoint detector. Sun \textit{et al.}~\cite{sun2019Perceive} propose a VPM to extract local features while simutaneously being aware of the visibility of each body part. However, VPM does not consider the deformation mentioned in DSR. Our method address the deformation by an explicit alignment step and gets rid of the time-consuming comparsion of each query-gallery pair during the inference stage.

\emph{Person search} considers the detection stage in raw video frames before reID. Current methods of person search can be divided into two categories. One integrates detection and reID into a unified framework~\cite{tong2017Joint,munjal2019Query,hao2017Neural,yichaoLearning}. Based on Faster-RCNN~\cite{ren2015faster}, OIM proposed by Xiao \textit{et al.}~\cite{tong2017Joint} inserts an additional feature extraction branch on top of the detector head, which is the first end-to-end learning framework for person search. Similar to NPSM of Liu \textit{et al.}~\cite{hao2017Neural}, Munjal \textit{et al.}~\cite{munjal2019Query} apply a query-guided method for person search while it is end-to-end optimized compared to NPSM. The other category solves detection and re-identification with two separate models, \emph{i.e.}, performing person search in a sequential manner~\cite{chen2018Person,xu2018Person,han2019Re}. Chen \textit{et al.}~\cite{chen2018Person} propose a mask-guided feature learning method to make the reID network focus more on foregrounds. Lan \textit{et al.}~\cite{xu2018Person} propose CLSA which utilizes multi-level features from reID network to solve the multi-scale matching problem. Han \textit{et al.}~\cite{han2019Re} claim that bounding box from detector might not be optimal for re-identification, they thus refine the detector results driven by re-identification loss.

Our work belongs to the first category of person search, integrating detection and reID into a unified framework. Different from existing person search works, this work further addresses the misalignment issue of detected bounding boxes and optimizes aligned part feature learning to achieve robust partial matching in person search.\vspace{-0.2cm}

\begin{figure*}[t]
\centering
\begin{center}
\includegraphics[width=0.96\textwidth]{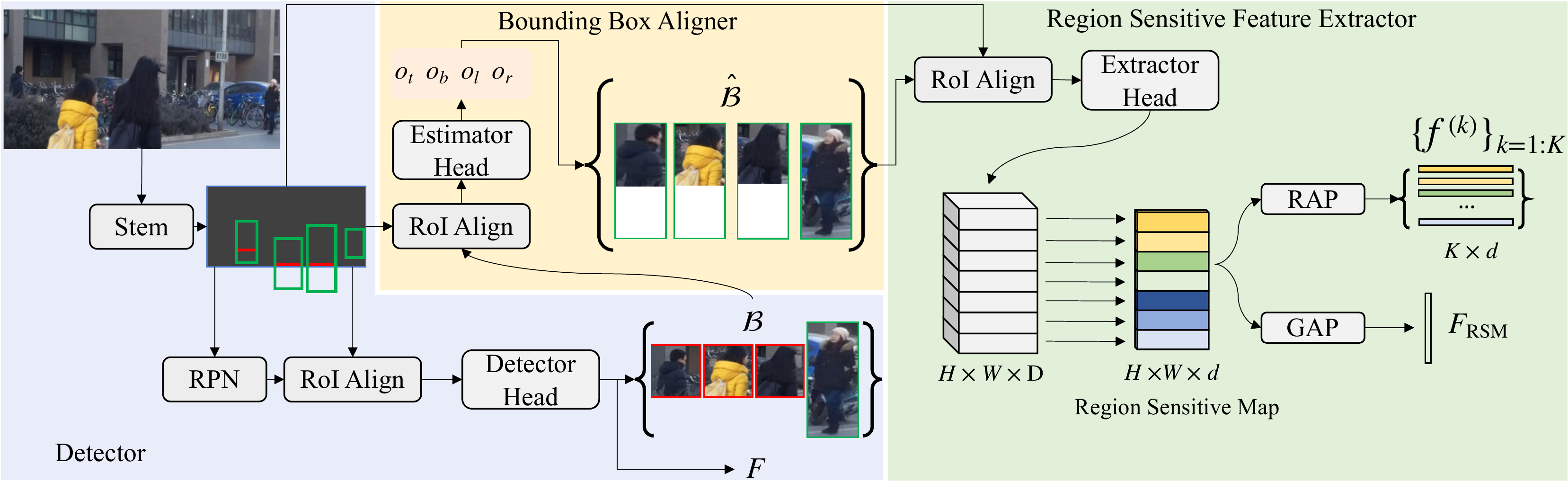}\\ 
\end{center}
\caption{Architecture of the proposed Align-to-Part Network (APNet), which consists of person detector, Bounding Box Aligner (BBA), and Region Sensitive Feature Extractor (RSFE). The detection branch returns bounding boxes $\mathcal{B}$ and global feature $F$. BBA refines detected boxes $\mathcal{B}$ to aligned ones, \emph{i.e.}, $\hat{\mathcal{B}}$. Red and green bounding boxes denotes detected boxes $\mathcal{B}$ and refined boxes $\hat{\mathcal{B}}$, respectively. RSFE extracts part features $\{f^{(k)}\}_{k=1:K}$ from refined boxes and alleviates negative effects of adjacent noises. RAP and GAP denote Regional Average Pooling and Global Average Pooling, respectively. $F_{\operatorname{RSM}}$ is the global feature from Region Sensitive Map, which is only used for training RSFE. }
\label{fig:adnet}
\vspace{-0.2cm}
\end{figure*}

\section{Problem Formulation}\label{sec:formulation}

Given a query person image $q$ and a gallery set $\mathcal G = \{g_i\}_{i=1:N}$ containing $N$ frames, person search aims at detecting a collection of person bounding boxes $\mathcal B = \{b_i\}_{i=1:M}$ from $\mathcal G$, then matching $q$ against the bounding boxes in $\mathcal B$. Suppose a person can be divided into $K$ body parts, we could represent the bounding box $b_i$ containing a complete person as a collection of $K$ parts,\emph{ i.e.}, $\mathcal P_i= \{p_i^{(k)}\}_{k=1:K}$, where $p_i^{(k)}$ denotes the $k$-th part.

With the above formulation, person detection is expected to return bounding boxes containing complete persons with $K$ parts. Person reID targets to extract a discriminative feature to identify the same identity of $q$ in $\mathcal B$. Most of existing works extract a global feature $F$ and performs reID by L2-distance. Due to the occlusions or background clutters, certain bounding boxes only contain portions of body parts, making the global feature degrades to a partial feature, \emph{i.e.}, the global feature is extracted from $l, l<K$ visible parts. Such partial features lead to inaccurate person matching when compared with global features.

Our solution is to introduce a part identifier $\operatorname E(\cdot)$ to identify visible parts in each detected bounding box, \emph{i.e.}, $\mathcal P_i = \operatorname E(b_i)$. With identified parts, the person image matching could be treated as a part feature matching task, where features on mutual visible parts of two bounding boxes are matched. The distance between $q$ and $b_i$ can be denoted as, \emph{i.e.},
\begin{equation}\label{eq:robust}
\operatorname{dist}_P(q, b_i)=\frac{\sum\limits_{k\in \mathcal P_q\cap \mathcal P_i}\operatorname D(f_q^{(k)}, f_i^{(k)})}{|\mathcal P_q\cap \mathcal P_i|},
\end{equation}
where $f_q^{(k)}$ and $f_i^{(k)}$ are features extracted from $k$-th part from $q$ and $b_i$, and $\operatorname D(\cdot)$ refers the L2-distance.

The training of our person search model should guarantee an accurate person detector, the reliable part identifier, and discriminative part features. We formulate our training objective as,
\begin{equation}\label{eq:loss}
\mathcal L= \mathcal L_D + \mathcal L_P + \sum\limits_{k=1}^{K} \mathcal L_{f^{(k)}},
\end{equation}
where $\mathcal L_D$ denotes loss of the detector, which is optimized with both bounding box locations and person reID. $\mathcal L_P$ denotes the loss of part identification. $\mathcal L_f^{(k)}$ evaluates the discriminative power of the $k$-th part feature, which can be implemented with reID loss. The following section presents details of our implementation to person detector, part identifier, part feature extraction, as well as the network optimization.

\section{Proposed Methods}

We propose the Align-to-Part Network (APNet) to implement the formulation in Sec.~\ref{sec:formulation}. The architecture of APNet is shown in Fig.~\ref{fig:adnet}. APNet consists of a person detector, Bounding Box Aligner (BBA), and the Region Sensitive Feature Extractor (RSFE), respectively. The following parts present details of those components.

The detector is built upon OIM~\cite{tong2017Joint}, which is an end-to-end person detector returning bounding boxes $\mathcal {B}$ as well as their corresponding global feature $F$. As shown in Fig.~\ref{fig:adnet}, the detector is trained with RPN loss ($\mathcal{L}_{\operatorname{rpncls}}$, $\mathcal{L}_{\operatorname{rpnreg}}$)~\cite{ren2015faster}, ROI Head loss ($\mathcal{L}_{\operatorname{cls}}$, $\mathcal{L}_{\operatorname{reg}}$), as well as reID loss. We denote the detector training loss as
\begin{equation}\label{eq:oimloss}
\mathcal L_D=  \mathcal{L}_{\operatorname{cls}}+\mathcal{L}_{\operatorname{reg}}+\mathcal{L}_{\operatorname{rpncls}}+\mathcal{L}_{\operatorname{rpnreg}}+\mathcal{L}_{\operatorname{ID}},
\end{equation}
where $\mathcal{L}_{\operatorname{cls}}$ and $\mathcal{L}_{\operatorname{reg}}$ denotes the person classification loss and bounding box regression loss in ROI Head. $\mathcal{L}_{\operatorname{rpncls}}$ and $\mathcal{L}_{\operatorname{rpnreg}}$ denote the  objectiveness loss and proposal regression loss in RPN. $\mathcal{L}_{\operatorname{ID}}$ is the reID loss computed on the global feature. We refer readers to OIM~\cite{tong2017Joint} for more details of the loss computation.

As shown in Fig.~\ref{fig:motivation}, a well trained detector may produce misaligned person bounding boxes. We hence design a part estimator with BBA to estimate visible body parts of each detected box. 

\begin{figure}[t]
\begin{center}
\centering
\includegraphics[width=0.9\linewidth]{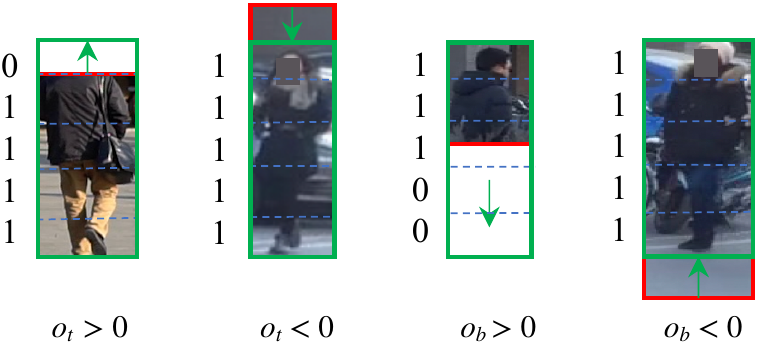}\\
\end{center}
\caption{Illustration of BBA for bounding box refinement. Red and green boxes refer to detected boxes and refined boxes, respectively. BBA predicts four bounding box offset values $o_t$, $o_b$, $o_l$, and $o_r$, which shift the detected bounding boxes to cover the estimated holistic body region. 0, 1 indicate the validity of each part. }
\label{fig:bve}
\vspace{-0.2cm}
\end{figure}

\subsection{Bounding Box Aligner}\label{sec:bba}

BBA implements part identifier $\operatorname E(\cdot)$ in Sec.~\ref{sec:formulation} to identify visible parts in each bounding box. This could be achieved with various methods, \emph{e.g.}, through segmenting person foregrounds~\cite{liFully} or estimating body keypoints~\cite{fang2017rmpe}. However, those methods require extra annotations and considerable computations. We implement BBA with a more efficient solution.

Because most pedestrians show upright posture in surveillance videos, aligned body parts can be extracted by dividing a holistic body region into horizontal and vertical stripes. This operation generates aligned part regions, \emph{e.g.}, top and bottom horizontal stripes correspond to head and foot, respectively. With this intuition, BBA first refines detected bounding boxes, then extracts horizontal and vertical stripes as body parts. We illustrate this procedure in Fig.~\ref{fig:bve}.

BBA estimates an offset vector $O=\{o_t, o_b, o_l, o_r\}$ to refine each detected bounding box to cover the holistic body region, where each offset value is in the range of $[-1, 1]$. The four offset values move the top, bottom, left, and right boundaries of each bounding box. Fig.~\ref{fig:bve} shows examples of moving the boundaries by $o_t$ and $o_b$. Suppose the coordinate of a bounding box $b$ is $\{x_{min}, y_{min}, x_{max}, y_{max}\}$, its coordinate after refinement with $O$ can be denoted as $\{\hat{x}_{min}, \hat{y}_{min}, \hat{x}_{max}, \hat{y}_{max}\}$, \emph{e.g.},
\begin{equation}\label{eq:bve}
\begin{split}
\hat{y}_{min} = y_{min}-\frac{h\cdot o_t}{1-o_t-o_b}\\
\hat{y}_{max} = y_{max}+\frac{h\cdot o_b}{1-o_t-o_b},
\end{split}
\end{equation}
where $h$ is the height of detected bounding box, computed as $y_{max}-y_{min}$. Similar computation can be applied to compute the $\hat{x}_{min}$ and $\hat{x}_{max}$ with $o_l$ and $o_r$.

We denote the refined bounding box as $\hat b$, which is used to extracted horizontal and vertical stripes as parts. As shown in Fig.~\ref{fig:bve}, refinements may introduce occluded parts and noises into $\hat b$. To extract $K$ horizontal stripes, we introduce a $K$-dim validity vector $v$ to record the visibility of each stripe. The $k$-th stripe is considered as valid, \emph{i.e.}, $v[k]=1$, if
\begin{equation}\label{eq:bve_identifier}
\lceil K\cdot \max(0,o_t) \rceil \leq k \leq K-\lfloor K\cdot \max(0,o_b) \rfloor\\.
\end{equation}
Similar computation is applied to extract valid vertical stripes. The final valid part collection $\mathcal{P}$ of each bounding box collects valid stripes.

As shown in Fig.~\ref{fig:adnet}, BBA predicts $O$ based on the bounding box feature extracted with ROIAlign. BBA can be trained by automatically generating training data. We first crop bounding boxes from frames by their groundtruth coordinates, which we denote as $\mathcal{B}_{gt}$. AlphaPose~\cite{fang2017rmpe} is utilized to estimate the keypoints of each bounding box, which provide cues about missing body parts. We then transform $\mathcal{B}_{gt}$ to $\hat{\mathcal{B}}_{gt}$ to cover the holistic body region.
Comparing $\mathcal{B}_{gt}$ and $\hat{\mathcal{B}}_{gt}$ generates the groundtruth offset labels $O_{gt}=\{o_{gt-t}, o_{gt-b}, o_{gt-l}, o_{gt-r}\}$. The training of BBA can thus be supervised by following loss
\begin{equation}\label{eq:bve_train}
\mathcal{L}_{P}=\sum\limits_{i\in\{t,b,l,r\}}\operatorname {smooth}_{l1}(o_i, o_{gt-i}),
\end{equation}
where $\operatorname {smooth}_{l1}$ computes differences between predicted offset values and the groundtruth values. More details of $\operatorname {smooth}_{l1}$ can be found in ~\cite{ren2015faster}.

\subsection{Region Sensitive Feature Extractor}\label{sec:rse}

The part collection $\mathcal{P}$ makes it possible to extract part features for partial matching. The following section shows our method to extract horizontal stripe features. Vertical stripe features can be extracted with similar way.

\textbf{Vanilla Part Feature Extractor:} Part features can be extracted by applying Region Average Pooling (RAP) on the feature map of video frame. As shown in Fig.~\ref{fig:adnet}, for a refined bounding box $\hat{b_i}$, we first extract its feature map from the frame feature map $\mathcal M$ with ROIAlign, which is then input into a convolutional block to generate a feature map $T\in \mathbb{R}^{H\times W\times D}$. We denote part features of $\hat {b_i}$ extracted by RAP as,
\begin{equation}\label{eq:bce_pcb_van}
\{\bar f_i^{(k)}\}_{k=1:l}=\operatorname{RAP}(T, \mathcal P_i), l=|\mathcal P_i|,
\end{equation}
where $l$ denotes the number of valid horizontal stripes of bounding box $\hat{b_i}$.

Part feature learning can be achieved by computing the reID loss on each valid part feature, \emph{i.e.},
\begin{equation}\label{eq:bba_rap}
\mathcal{L}_{f}^{(k)}=\mathcal{L}_{\operatorname{ID}}(\bar {f}^{(k)}, y),
\end{equation}
where $\mathcal{L}_{\operatorname{ID}}$ refers to reID loss which is implemented with OIM loss~\cite{tong2017Joint}. $y$ is the ground truth person ID label.

\begin{figure}[t]
\begin{center}
\centering
\includegraphics[width=0.45\textwidth]{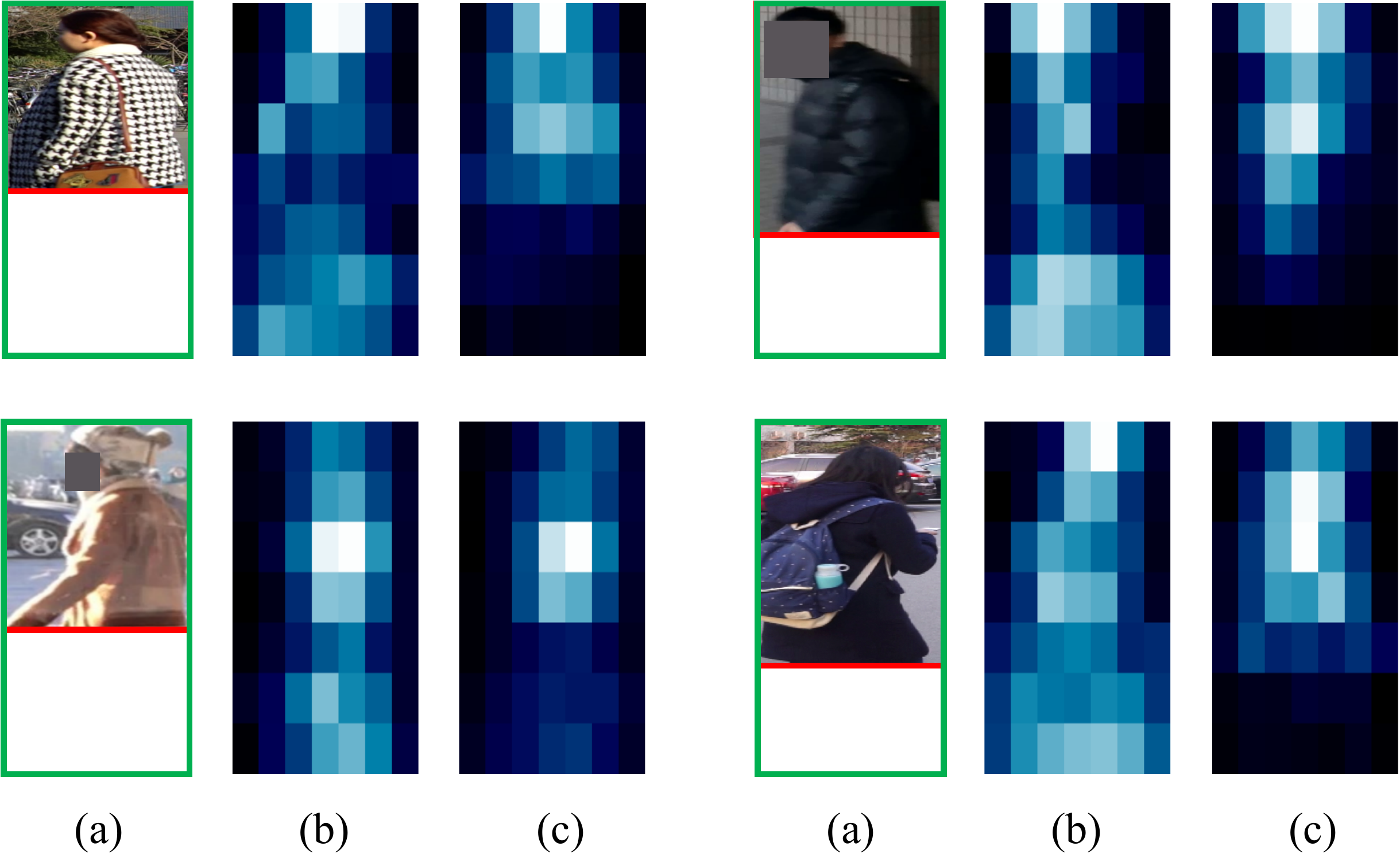} 
\end{center}
\caption{Illustration of detected boxes (red) and refined boxes (green), as well as their feature maps. (b) and (c) show feature maps of refined boxes trained with vanilla part feature extractor and our RSFE. It is clear that, RSFE depresses noises on noisy and invalid regions.}
\label{fig:psmap_vis}
\vspace{-0.4cm}
\end{figure}

Fig.~\ref{fig:psmap_vis} (b) shows feature maps of aligned bounding boxes trained by the above feature extractor. It is clear that, noisy or invisible regions still have strong responses. This could be because the training procedure in Eq.~\eqref{eq:bba_rap} focuses on visible body parts, and can not tune the features on noisy parts. Because of enlarged receptive fields of neurons in feature extraction CNN layer, the strong CNN activations on noisy or occluded parts would affect features of their adjacent valid parts. This may degrades the validity of the above feature extractors. Therefore, part feature extractors robust to occlusions and noises are required.

\textbf{Feature Extraction with Region Sensitive Map:} Our RSFE introduces a Region Sensitive Map (RSM) to suppress responses of invalid regions. To tune the responses of each feature stripe, we generate the RSM $M\in \mathbb{R}^{H\times W\times d} $ by applying 1 $\times$ 1 convolution on each horizontal feature stripe $T[i]\in\mathbb{R}^{W\times D}, i=1:H$. The computation of $M[i]\in\mathbb{R}^{W\times d}$ can be denoted as,
\begin{equation}\label{eq:bce_rsm}
M[i]= \operatorname{conv^{(i)}_{1\times 1}}(T[i]), i=1:H,
\end{equation}
where $\operatorname{conv^{(i)}_{1\times 1}}$ refers to the $i$-th convolution kernel. Different from a global $1\times 1 $ convolution, $H$ convolutional kernels in Eq.~\eqref{eq:bce_rsm} do not share parameters to deal with complicated occlusions on different spatial locations.

To suppress noisy regions, we supervise ${M}$ with a global feature. The loss can be computed by applying Average Pooling on $M$ and supervise the resulted feature $F_{\operatorname{RSM}}$, \emph{i.e.},
\begin{equation}\label{eq:bce_rsm_sup}
\mathcal{L}_{\operatorname{RSM}}=\mathcal{L}_{\operatorname{ID}}(F_{\operatorname{RSM}}, y).
\end{equation}
This loss enforces the network to depress activations on noisy regions to generate a better $F_{\operatorname{RSM}}$, which is then used for stripe feature extraction. Fig.~\ref{fig:psmap_vis} (c) shows feature maps after training with Eq.~\eqref{eq:bce_rsm_sup}, where invalid regions are effectively suppressed.

From $M$, the stripe features can be extracted by RAP as,
\begin{equation}\label{eq:bce_rsm_rap}
\{f_i^{(k)}\}_{k=1:l}=\operatorname{RAP}(M,\mathcal{P}_i), l=|\mathcal P_i|,
\end{equation}
where each part feature can be trained with part feature loss in Eq.~\eqref{eq:bce_pcb_van}.

Eq.~\eqref{eq:bce_rsm} applies different $1\times1$ convolution kernels on spatial locations of $T$. This enables more specific refinements for each part feature. Besides that, different spatial locations show varied probabilities of occlusion, \emph{e.g.}, the foot area is more likely to be occluded. This property makes Eq.~\eqref{eq:bce_rsm} more effective in depressing occlusions and noises than learning a global $1\times1$ convolution kernel.

The methods above extracts features from horizontal stripes. Similar procedure can be applied to extract features from vertical stripes. Given a query person image $q$ and a detected gallery bounding boxes $b_i$ to be compared, we utilize both global and part features for person reID. The part feature distance can be computed with Eq.~\eqref{eq:robust}. We also utilize the global feature $F$ from detector branch for distance computation. The overall distance between $q$ and $b_i$ can be computed as:
\begin{equation}\label{eq:global_feature}
\operatorname{dist}(q, b_i) = \operatorname{dist}_P(q, b_i) + \lambda \cdot\operatorname{D}(F_q, F_i),
\end{equation}
where $F_q$ and $F_i$ correspond to the global features of $q$ and $b_i$ respectively. We set $\lambda$ to 1 in the following experiments.

\section{LSPS Dataset}

\subsection{Previous Datasets}
\textbf{CUHK-SYSU}~\cite{tong2017Joint} consists of 5,694 frames captured from movie snapshots, and 12,490 frames from street snap. Both bounding boxes and identities are labeled manually. The dataset provides 8,432 labeled identities, and 23,430 labeled bounding boxes. 96,143 bounding boxes are provided in total. 11,206 frames with 5,532 identities make up the training set. CUHK-SYSU does not annotate bounding boxes with partial bodies.

\textbf{PRW}~\cite{zheng2017Person} is captured with six cameras deployed at a campus, containing 11,816 frames with 932 identities in total. 5,134 frames are selected as training set with 432 identities, while the rest 6,112 frames make up the test set. A total of 34,304 bounding boxes are annotated with identity. Similar to CUHK-SYSU, bounding box locations in PRW are also manually labeled. PRW includes some misaligned bounding boxes in both queries and galleries.

\begin{table}
\begin{center}
\resizebox{0.99\linewidth}{!}{
\small
\begin{tabular}{p{1.8cm}|p{1.8cm}<{\centering}|p{1.5cm}<{\centering}|p{1.5cm}<{\centering}}
\hline
Dataset& LSPS & PRW~\cite{zheng2017Person}& CUHK~\cite{tong2017Joint} \\
\hline
frames& 51,836& 11,816& 18,184 \\
identities & 4,067& 932& 8,432 \\
anno. boxes&  60,433& 34,304& 23,430 \\
cameras& 17& 6& - \\
detector & Faster-RCNN& Hand& Hand\\
inc.query& $\sim$60\%&$\sim$6\%& $\sim$0\%\\
\hline
\end{tabular}}
\end{center}

\caption{Comparison between LSPS and the other two person search datasets. ``detector'' refers to the way to obtain the ground truth of bounding box location. ``inc.query'' means the percentage of query bounding boxes with partial body.} \label{statistics}
\vspace{-0.3cm}
\end{table}

\subsection{Description to LSPS}

\begin{figure}[t]

\begin{center}
\includegraphics[width=0.45\textwidth]{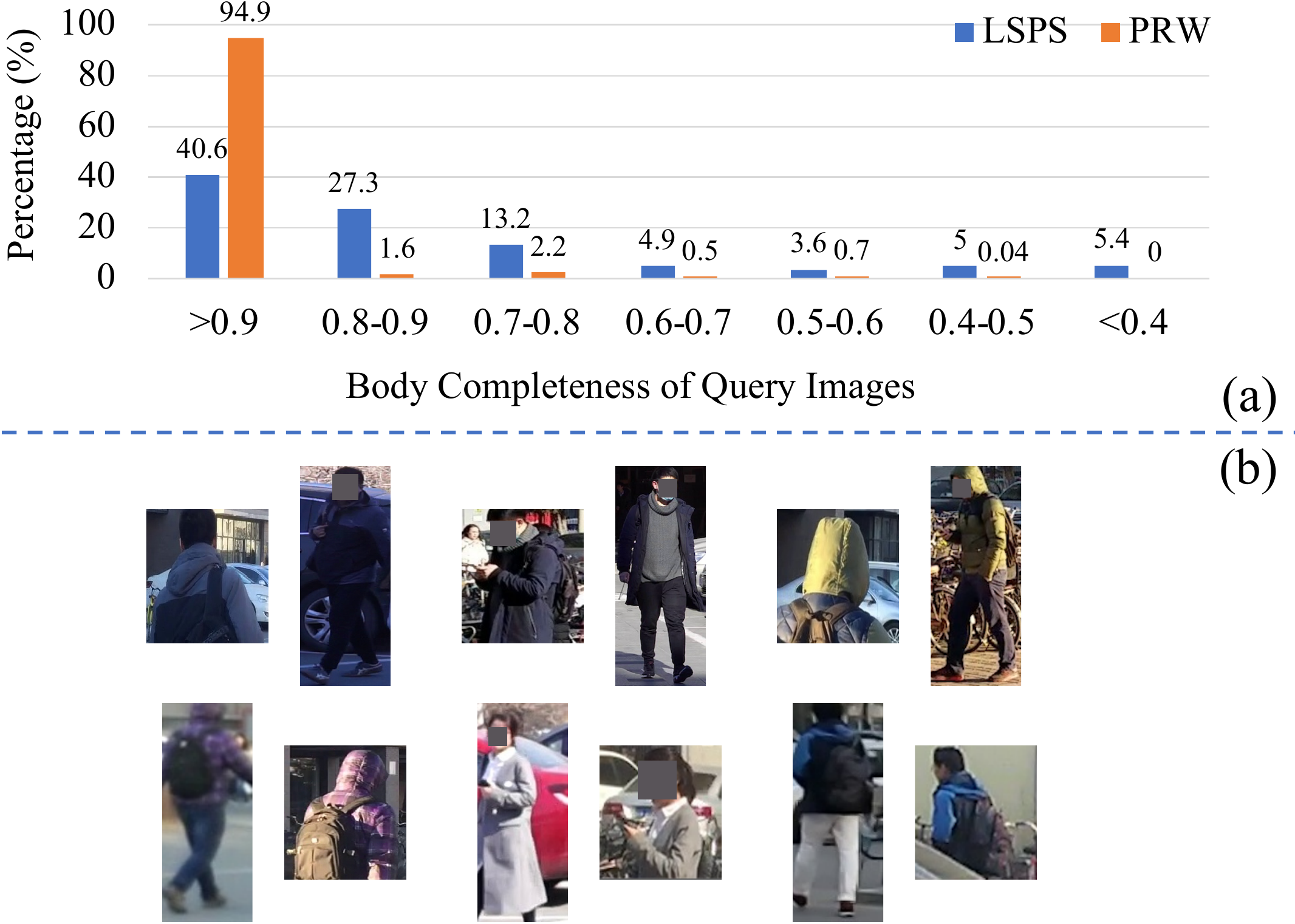} 
\end{center}\caption{(a) compares the body completeness in queries between LSPS and PRW. (b) illustrates several query and gallery bounding boxes. It can be observed that, misalignment occurs in both query and gallery bounding boxes.}
\label{fig:rwpid}
\end{figure}

This paper contributes a new \textit{Large-Scale dataset for Person Search in the wild} (LSPS), which shows the following characteristics:

\textit{Complex scene and appearance variations:} The video frames are collected from 17 cameras, deployed at both indoor and outdoor scenes. Different cameras exhibit different backgrounds, viewpoints, view fields, illuminations, pedestrian densities, \emph{etc}. Those factors lead to substantial appearance variances for the same person. Meanwhile, due to limited view fields of each camera and high person density, lots of pedestrians are occluded, leading to bounding boxed covering partial body region. LSPS includes partial bounding boxes into both query and gallery sets. Fig.~\ref{fig:rwpid} (a) compares the body completeness in queries between LSPS and PRW. It is clear that, LSPS has a substantially larger number of incomplete query bounding boxes. Compared with existing person search datasets, LSPS presents more complex scenes and appearance variations.

\textit{Larger scale:} Different with previous benchmarks, where bounding boxes are manually labeled. LSPS utilizes bounding boxes detected by Faster-RCNN~\cite{ren2015faster}, based on which we manage to collect a dataset with larger number of bounding boxes. We show the comparison between LSPS and the other two datasets in Table~\ref{statistics}. In LSPS, a total number of 51,836 frames are collected, from which 60,433 bounding boxes and 4,067 identities are annotated. Table~\ref{statistics} shows that, LSPS presents the larger number of frames, annotated person bounding boxes, cameras, respectively. Moreover, about 60\% queries in LSPS cover partial body. Larger scale and incomplete query bounding boxes make LSPS a more challenging and realistic dataset than the others in Table~\ref{statistics}.

\subsection{Evaluation Protocol}
LSPS is splitted into a training set with 18,154 frames and a test set with 33,673 frames. The training and test set contain 1,041 and 3,026 identities, respectively. Statistics of the training and test set are summarized in Table~\ref{statistics_traintest}. We utilize mean Average Precision (mAP) and rank-1 accuracy as evaluation metrics, which are widely adopted in person reID ~\cite{tong2017Joint,munjal2019Query,chen2018Person,xu2018Person,han2019Re}. In person search, a retrieved bounding box is considered as true positive if it shares the same ID label with query and it has the overlap rate larger than 0.5 with the ground truth bounding box. Therefore, both mAP and rank-1 accuracy in person search are affected by the detector performance.

\begin{table}[t]
\begin{center}
\small
\begin{tabular}{p{1.cm}|p{1cm}<{\centering}|p{1.3cm}<{\centering}| p{1.1cm}<{\centering}|p{1.6cm}<{\centering}}
\hline
Split& frames & identities& boxes& anno. boxes\\
\hline
Training& 18,163& 1,041& 71,563& 18,928 \\
Test& 33,673& 3,026& 116,170& 41,505 \\

\hline
\end{tabular}
\end{center}
\caption{Statistics of training/test sets on LSPS. ``anno. boxes'' refers bounding boxes annotated with person ID. ``boxes'' denotes the number of provided bounding boxes.} \label{statistics_traintest}
\vspace{-0.32cm}
\end{table}

\section{Experiment}
The following parts conduct experiments on CUHK-SYSU~\cite{tong2017Joint}, PRW~\cite{zheng2017Person} and the new LSPS. 

\subsection{Implementation Details}
Our APNet is implemented based on OIM~\cite{tong2017Joint}. We use the ResNet50~\cite{he2016deep} initialized with ImageNet pretrained model as the backbone. The backbone adopts similar setting with the one in~\cite{munjal2019Query} and it uses RoIAlign~\cite{he2017Mask} rather than RoIPooling~\cite{ren2015faster}. For all experiments, we train the APNet with SGD optimizer. We divide the training into two stages, \textit{i.e.}, the first stage trains the detector branch with $\mathcal{L}_D$. The second stage fixes parameters of the detector and trains BBA and RSFE with $\mathcal{L}_P$ and $\mathcal{L}_f$, respectively. We set $K$ to 7 for horizontal stripes. For PRW and CUHK-SYSU, we do not use features from vertical stripes since the misalignment in these two datasets mainly occurs in vertical direction.

On CUHK-SYSU, the first stage lasts 40k iterations. We set the initial learning rate to 1e-3 and decay it by 0.1 at 30k. RSFE is trained for 40k with learning rate as 1e-3, decayed at 30k. On PRW, the first stage lasts 80k. We set learning rate to 1e-4, decayed at 60k. RSFE is trained for 40k with learning rate as 1e-3, decayed at 30k. On LSPS, the first stage lasts 120k, with learning rate as 1e-4 and is decayed by 0.1 at 80k and 100k, respectively. RSFE is trained for 60k with learning rate as 1e-3, decayed at 30k and 50k, respectively. BBA on three datases are trained for 30k with learning rate as 1e-3, and is decayed at 20k.

\subsection{Ablation Study}

\begin{table}[t]
\begin{center}
\resizebox{.9\linewidth}{!}{
\tiny
\begin{tabular}{p{1cm}<{\centering}|p{1.8cm}<{\centering}|p{1.8cm}<{\centering}|p{1.8cm}<{\centering}}
\hline
\multicolumn{2}{c|}{\multirow{1}*{Methods}} &

\multicolumn{1}{c|}{\multirow{2}*{mAP}} & \multicolumn{1}{c}{\multirow{2}*{rank-1}}\\
\cline {1-2}

\multicolumn{1}{l|}{}&\multicolumn{1}{l|}{feature}&\multicolumn{1}{l|}{}&\multicolumn{1}{l}{}\\
\cline{1-4}

\multicolumn{1}{l|}{\multirow{1}*{upper bound}}&
\multicolumn{1}{l|}{global} & \multicolumn{1}{c|}{53.5}& \multicolumn{1}{c}{74.6}\\
\cline{1-4}

\multicolumn{1}{l|}{\multirow{1}*{global}}&
\multicolumn{1}{l|}{global} & \multicolumn{1}{c|}{13.6}& \multicolumn{1}{c}{30.4}\\

\cline {2-2}
\multicolumn{1}{l|}{\multirow{1}*{matching}}&
\multicolumn{1}{l|}{stripe} & \multicolumn{1}{c|}{23.1}& \multicolumn{1}{c}{52.5}\\
\cline {1-4}
\multicolumn{1}{l|}{\multirow{1}*{partial}}&
\multicolumn{1}{l|}{{valid region}} & \multicolumn{1}{c|}{23.9}& \multicolumn{1}{c}{46.7}\\
\cline {2-2}
\multicolumn{1}{l|}{\multirow{1}*{matching}}&
\multicolumn{1}{l|}{\textit{mutual} region} & \multicolumn{1}{c|}{21.4}& \multicolumn{1}{c}{48.6}\\
\cline {2-2}
\multicolumn{1}{c|}{}&
\multicolumn{1}{l|}{\textit{mutual} stripe} & \multicolumn{1}{c|}{\textbf{30.9}}& \multicolumn{1}{c}{\textbf{70.0}}\\

\hline

\end{tabular}}
\end{center}
\caption{Validity of partial matching on Market1501 modified by adding artificial occlusions. ``upper bound'' refers to performance of global feature on original Market-1501. Other methods are tested on modified Market-1501. Valid region refers to common visible region of tow matched bounding boxes.} \label{exp:abla_global}
\vspace{-0.2cm}
\end{table}

\textbf{Validity of Partial Matching.} APNet addresses misalignment issue with partial matching based on stripe features. This part demonstrates the validity of this strategy. We first modify the Market-1501~\cite{zheng2015scalable} by adding random occlusions to query and gallery images. Based on this modified dataset, we compare different feature matching strategies and summarize the results in Table~\ref{exp:abla_global}, where ``global matching'' does not differentiate occlusions in feature extraction and ``partial matching'' extracts features from valid regions.

Table~\ref{exp:abla_global} shows that, occlusion is harmful for reID, \emph{e.g.}, degrades the mAP of global feature from 53.5\% to 13.6\%. Extracting stripe features from the entire bounding box boosts the reID performance. This shows the validity of part feature. Because partial matching extracts features from visible regions, it achieves better performance than global matching. Among three features used in partial matching, features extracted from mutual stripes achieve the best performance. It outperforms the feature from mutual region by large margins, \emph{e.g.}, 30.9\% vs. 21.4\% in mAP. It also significantly outperforms the global feature in global matching. The above experiments hence shows the validity of the idea illustrated in Fig.~\ref{fig:motivation}, where features on the mutual visible regions are matched for similarity computation.

\begin{table}[t]

\centering
\begin{center}
\setlength{\tabcolsep}{0.3cm}
\resizebox{1.\linewidth}{!}{
\normalsize
\begin{tabular}{p{2.2cm}|p{1.2cm}<{\centering}|p{1.35cm}<{\centering}|p{1.2cm}<{\centering}|p{1.35cm}<{\centering}}
\hline
\multicolumn{1}{l|}{\multirow{2}*{Features}} &

\multicolumn{2}{c|}{PRW} & \multicolumn{2}{c}{LSPS}\\
\cline {2-5}
\multicolumn{1}{c|}{}& mAP & rank-1 & mAP & rank-1\\
\hline

global& 34.2 & 75.8 & 14.4 & 47.7 \\
stripe& 39.1& 79.1 & 13.0& 41.6\\
VPM~\cite{sun2019Perceive} & 40.0& 80.2& 16.0& 49.5\\
stripe (BBA) &  40.8& 81.0& 16.4& 50.2\\
RSFE (BBA)&  \textbf{41.9}& \textbf{81.4}& \textbf{17.1}& \textbf{51.6}\\

\hline
\end{tabular}}
\end{center}
\caption{Validity of BBA and RSFE in stripe feature extraction. Global refers to global feature extracted from detected bounding box. ``stripe (BBA)'' denotes extracting stripe features with vanilla part feature extractor based on BBA output. ``RSFE (BBA)'' denotes stripe feature extracted by RSFE.}\label{exp:abla}
\vspace{-0.2cm}
\end{table}

\begin{figure}[t]
\begin{center} 
\centering
\includegraphics[width=0.45\textwidth]{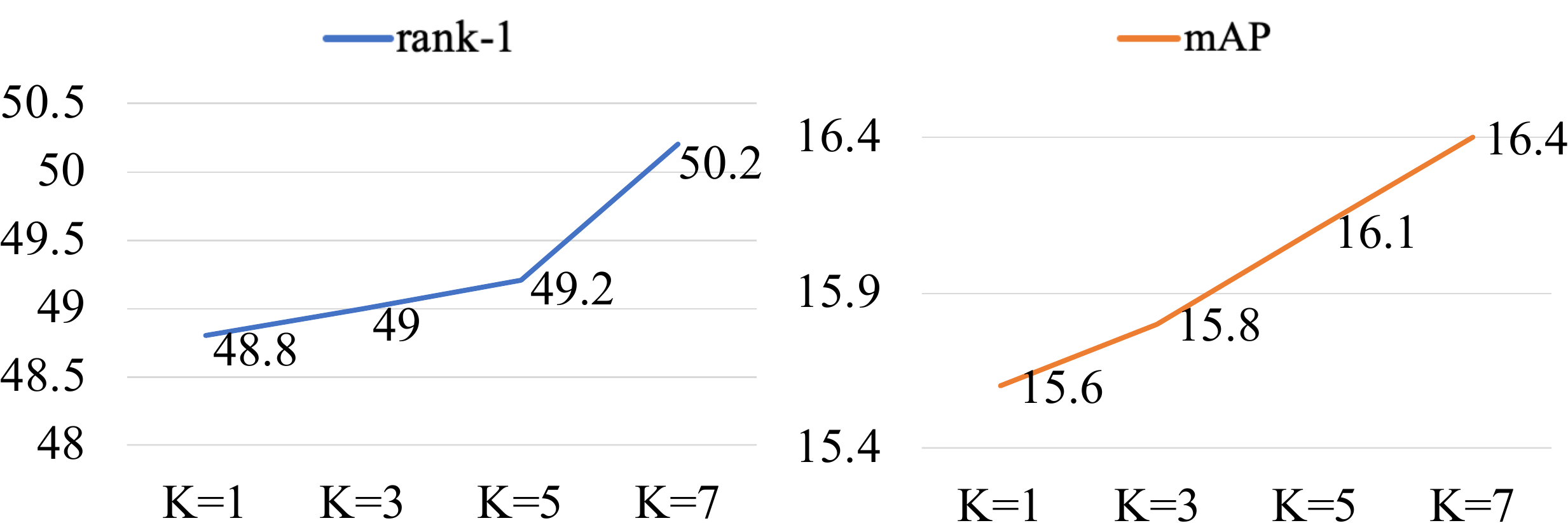} 
\end{center}
\caption{Performance on LSPS with different stripe number $K$. }
\label{fig:nstripe}
\vspace{-0.2cm}
\end{figure}

\textbf{Validity of BBA:} BBA performs bounding box refinement and valid part estimation. We proceed to verify the validity of BBA in person search. Table~\ref{exp:abla} shows the performance of global feature and stripe feature before applying BBA. It is clear that, extracting valid stripe features from BBA refined bounding boxes, substantially boosts the reID performance. For instance, ``stripe (BBA)'' achieves mAP of 40.8\% on PRW, better than the 34.2\% and 39.1\% of original global feature and stripe feature, respectively. We also show performance of the reproduced VPM~\cite{sun2019Perceive}, which is a recent method for partial matching. As shown in Table~\ref{exp:abla}, stripe feature extracted with BBA performs better than VPM on both PRW and LSPS. This experiment hence shows the validity of BBA in bounding box refinement and valid part extraction.

\textbf{Validity of RSFE:} RSFE is designed to alleviate the negative effects of noisy regions illustrated in Fig.~\ref{fig:psmap_vis} (b). It is also designed to refine each part feature. Table~\ref{exp:abla} compares the stripe feature extracted by RSFE, \emph{i.e.}, ``RSFE (BBA)'' against the one extracted by vanilla part feature extractor, i.e., ``stripe (BBA)''. It can be observed that, RSFE is important in boosting the stripe feature performance. For example, RSFE boosts the mAP by 1.1\% and 0.7\% on PRW and LSPS, respectively. The combination of BBA and RSFE achieves the best performance and outperforms the VPM by large margins. More comparisons with person search works will be presented in the next section.

\textbf{Effects of part number $K$:} We proceed to study the influence of part number $K$, and summarize the results in Fig.~\ref{fig:nstripe}. $K=1$ degrades the partial feature to global feature. The maximum horizontal stripe number $K$ equals to the height of $T$, \emph{i.e.}, 7 in our experiment. The figure shows that, stripes with finer scale help to boost the performance. This might due to the fact that larger $K$ improves the feature resolution and robustness to occlusions and noises.

\begin{figure}[t]
\begin{center}
\centering
\includegraphics[width=0.45\textwidth]{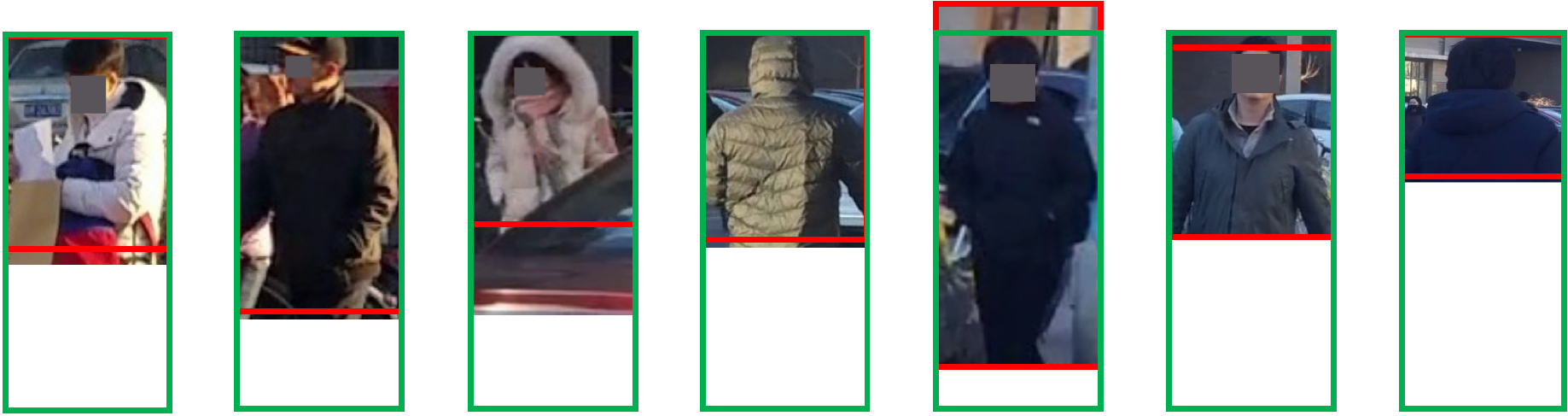} 
\end{center}
\caption{Visualization of refined bounding boxes by BBA. Red and green boxes denote bounding boxes before and after refinement. BBA effectively estimates the holistic body region to eliminate misalignment errors. }
\label{fig:padreg_vis}
\end{figure}

\textbf{Discussions:}. To show the effects of BBA in bounding boxes refinement, we visualize bounding boxes before and after refinement in Fig.~\ref{fig:padreg_vis}. The results show that, BBA effectively shifts the original bounding boxes to cover holistic body regions. This procedure eliminates misalignment errors and guarantees aligned stripe feature extraction.

Besides the detector branch based on baseline OIM, APNet introduces additional BBA and RSFE modules. We compare the parameter and computational complexity between OIM and APNet in Table~\ref{exp:overhead}. The comparison shows that APNet achieves promising performance with comparable speed with baseline OIM, \emph{e.g.}, 0.397 TFLOPs of APNet vs. 0.383 TFOLPs of OIM. Although the BBA and RSFE modules bring more parameters to APNet, they do not degrade its computational speed substantially. APNet is likely to be faster than person search works that treat detection and reID in separated steps, and is faster than some works like QEEPS~\cite{munjal2019Query}, which compares each query-gallery pair for person search.

\begin{table}[t]
\begin{center}
\small
\resizebox{0.99\linewidth}{!}{
\begin{tabular}{p{1.8cm}|p{2cm}<{\centering} |p{2.cm}<{\centering}| p{1.2cm}<{\centering}}
\hline
Methods& \# params (\textit{M}) & speed (sec.)& TFLOPs \\

\hline
OIM& 36& 0.254& 0.383\\
APNet& 67& 0.256& 0.397 \\

\hline
\end{tabular}}
\end{center}
\caption{Comparison of parameter and computational complexity between APNet and baseline OIM~\cite{tong2017Joint}. Speed and TFLOPS are measured on the NVIDIA 2080Ti GPU. } \label{exp:overhead}
\end{table}


\subsection{Comparison with recent works}

\begin{table}[t]
\centering
\begin{center}
\Large
\setlength{\tabcolsep}{0.3cm}
\resizebox{0.99\linewidth}{!}{
\begin{tabular}{p{2.1cm}|p{1.7cm}<{\centering}|p{1.5cm}<{\centering}|p{1.5cm}<{\centering}|p{1.5cm}<{\centering}|p{1.5cm}<{\centering}}
\hline

\multicolumn{1}{c|}{\multirow{2}*{Methods}} & \multicolumn{1}{c|}{\multirow{2}*{Reference}}& \multicolumn{2}{c|}{\bfseries{CUHK-SYSU}} & \multicolumn{2}{c}{\bfseries{PRW}} \\
\cline {3-6}
\multicolumn{1}{c|}{}& \multicolumn{1}{c|}{}& mAP & rank-1 & mAP & rank-1 \\
\hline

OIM~\cite{tong2017Joint}& CVPR2017 & 75.5& 78.7& 21.3& 49.9\\
NPSM~\cite{hao2017Neural}& ICCV2017 & 77.9& 81.2& 24.2& 53.1\\
CLSA~\cite{xu2018Person}& ECCV2018& 87.2& 88.5& 38.7& 65.0\\
MGTS~\cite{chen2018Person}& ECCV2018& 83.0& 83.7& 32.6& 72.1\\
CGPS~\cite{yichaoLearning}& CVPR2019& 84.1& 86.5& 33.4& 73.6\\
QEEPS~\cite{munjal2019Query}& CVPR2019& 88.9& 89.1& 37.1& 76.7\\
RDLR~\cite{han2019Re}& ICCV2019& \textbf{93.0}& \textbf{94.2}& \textbf{42.9}& 72.1\\
\hline
APNet& & 88.9& 89.3& 41.9& \textbf{81.4}\\
\hline
\end{tabular}}
\end{center}
\caption{Comparison with recent works on CUHK-SYSU and PRW, respectively. }\label{sota}
\vspace{-0.4cm}
\end{table}

\textbf{CUHK-SYSU.} We experiment on CUHK-SYSU with gallery size 100. APNet achieves the rank-1 accuracy of 89.3\% and mAP of 88.9\%, outperforming most of recent works. Note that RDLR~\cite{han2019Re} uses a stronger backbone ResNet50-FPN as well as ranking-based loss. Methods like CLSA~\cite{xu2018Person}, MGTS~\cite{chen2018Person}, and RDLR solve detection and reID with two separate models, which are expensive in both computation and storage. Compared with those works, our APNet is a unified model and is likely to present better computational and memory efficiencies.

\textbf{PRW.} On PRW, APNet achieves competitive performance, \emph{e.g.}, 81.4\% in rank-1 accuracy and 41.9\% in mAP, outperforming most of recent works. APNet also significantly outperforms the RDLR~\cite{han2019Re} in rank-1 accuracy with a weaker backbone, \emph{i.e.}, by 9.3\% in rank-1 accuracy. Since certain query images in PRW cover partial body parts, APNet exhibits more substantial advantages with partial matching. It also outperforms CGPS~\cite{yichaoLearning} and QEEPS~\cite{munjal2019Query} by 7.8\% and 4.7\% in rank-1, respectively. Note that, CGPS~\cite{yichaoLearning} and QEEPS~\cite{munjal2019Query} input each query-gallery pair into CNN for similarity computation, hence also present inferior retrieval efficiency compared with APNet.

\begin{table}[t]
\centering
\begin{center}

\resizebox{0.9\linewidth}{!}{
\Large
\begin{tabular}{p{1.4cm}|p{2cm}<{\centering}|p{2.5cm}<{\centering}|p{2cm}<{\centering}|p{2cm}<{\centering}}
\hline

{\bfseries LSPS}& OIM~\cite{tong2017Joint} &VPM~\cite{sun2019Perceive}& APNet& APNet+v \\
\hline
mAP& 14.4&  16.0& 17.1&\textbf{18.8}\\
rank-1& 47.7&  49.5& 51.6&\textbf{55.7}\\

\hline
\end{tabular}}
\end{center}
\caption{Comparisons with recent works on LSPS. APNet denotes our method. APNet+v considers extra vertical stripe features.}\label{sota_rwpid}
\vspace{-0.3cm}
\end{table}

\textbf{LSPS.} We finally present experimental results on LSPS. OIM~\cite{tong2017Joint} is compared as the baseline. APNet brings 3.9\% rank-1 improvement over the baseline, achieving 51.6\% and 17.1\% in rank-1 and mAP, respectively. APNet also outperforms the recent VPM~\cite{sun2019Perceive}, which solves the partial reID problem. To consider the misalignment in horizontal direction, we further apply additional vertical stripe features. The corresponding method, \emph{i.e.}, APNet+v achieves the best performance, achieving 55.7\% and 18.8\% in rank-1 and mAP, respectively. It is also clear that, the performance of APNet on LSPS is substantially lower than the ones on PRW and CUHK-SYSU, indicating the challenges of LSPS.
\vspace{-0.5cm}
\section{Conclusion}
This paper proposes an APNet to solve the bounding box misalignment issues in person search. APNet refines detected boxes to cover the estimated holistic body regions, meanwhile extracting part features from visible body parts. This formulates reID as a partial feature matching procedure, where valid part features are selected for similarity computation, while part features on occluded or noisy regions are discarded. This paper also contributes a LSPS dataset, which is by far the largest and the most challenging dataset for person search. Experiments show that APNet brings considerable performance improvement on LSPS. Moreover, it achieves competitive performance on existing person search benchmarks like CUHK-SYSU and PRW.

\noindent \textbf{Acknowledgments} This work is supported in part by The National Key Research and Development Program of China under Grant No. 2018YFE0118400, in part by Beijing Natural Science Foundation under Grant No. JQ18012, in part by Natural Science Foundation of China under Grant No. 61936011, 61425025, 61620106009, 61572050, 91538111.

{\small
\bibliographystyle{ieee_fullname}
\bibliography{egbib}
}

\end{document}